\documentclass{style/hcrg-nasa}

\issuer{Harvard Computational Robotics Group}
\edition{Preprint}

\usepackage[utf8]{inputenc}
\usepackage[T1]{fontenc}
\usepackage{url}
\usepackage{booktabs}
\usepackage{amsmath}
\usepackage{amssymb}
\usepackage{amsfonts}
\usepackage{array}
\usepackage{tabularx}
\usepackage{graphicx}
\usepackage{xspace}
\usepackage{float}
\usepackage{hyperref}

\usepackage{microtype}
\usepackage{xcolor}
\usepackage{tikz}
\usetikzlibrary{arrows.meta,backgrounds,calc,positioning}

\definecolor{textgray}{HTML}{555555}
\definecolor{memcolor}{HTML}{2B5B84}
\definecolor{plancolor}{HTML}{B85B14}
\definecolor{vlacolor}{HTML}{2D8A4E}
\definecolor{classicalcolor}{HTML}{6B4C9A}
\definecolor{stagefill}{HTML}{F4F4F6}

\newcommand{\method}{OCC4M\xspace}
\newcommand{\framesamp}{FrameSamp\xspace}

\newenvironment{ack}{\section*{Acknowledgments}}{}

\title{
OCC4M: Object-Centric 4D Memory for Spatiotemporal Reasoning
in Long-Horizon Manipulation
}
\author[1*]{Jack B. Jedlicki}
\author[1,2*]{Tanguy Dieudonné}
\author[1]{Heng Yang}
\affiliation[1]{Harvard University}
\affiliation[2]{ETH Zürich}
\contribution[*]{Equal contribution.\\
\texttt{jackbjed@g.harvard.edu}\quad\texttt{tdieudonne@ethz.ch}}

\abstract{
Long-horizon manipulation often requires reasoning about state absent from the
current view, such as a vanished object's location, temporal identity, or the
contents of a shuffled container. We present \method (``Occam''), an
object-centric 4D memory that maintains persistent tracks in a shared world
frame and explicitly represents temporal, motion, and containment relations.
A vision-language model (VLM) queries this structured memory to select actionable targets for
history-free low-level execution. Across seven simulation conditions and 350 episodes, \method achieves 96.6\% memory success and 88.9\%
end-to-end success, versus 54.6\% and 57.7\% for \framesamp, a raw-history VLM baseline using Gemini 3.7 Flash with the complete observation history and the same executor. In a
controlled viewpoint-transfer test, \method maintains 100\% memory and 98\%
end-to-end success after a viewpoint change, while full-history \framesamp falls to near-zero success. On 20 fixed-camera Franka episodes, \method reaches 85\% joint memory accuracy, versus at most 30\% for \framesamp across context sizes from $K=16$ to the complete history, and completes 45\% of full two-stage tasks. These results support explicit object-centric memory for persistent spatiotemporal reasoning in long-horizon manipulation. Qualitative videos are available at \url{https://occ4m-sup.github.io/occ4m-supplementary/}.
}

\begin{document}
\maketitle

\section{Introduction}

Consider a robot asked to place a mug where a kettle used to be. The kettle is
gone, and the robot may now view the counter from a different base pose. The
target is a historical place: the robot must recover its location and express
it in the current view. Similar dependencies arise for the first of several
identical objects, an object's pre-motion pose, the contents of a shuffled
container, or a workspace the robot has driven away from. These tasks require
memory that preserves both \emph{what a reference denotes} and \emph{where to act}.

History-aware controllers encode past observations inside a learned policy
\citep{rdpg,gmp,hamlet,robomme}. Multimodal models can instead reason over sampled
frames \citep{gemini-er,gemini25}, identifying the relevant event and recovering
a target at query time. Retrieval-based methods such as MemER learn to select
relevant keyframes for hierarchical robot control \citep{memer}. These approaches
address historical context, but recovering an earlier observation and expressing
its target in a new viewpoint are distinct requirements. Additional frames do
not themselves specify a persistent metric reference frame.

Object-centric maps and scene graphs provide explicit spatial state
\citep{conceptgraphs,hydra,qi2026compose}; 3D-Mem combines multi-view memory snapshots with
exploration and retrieval \citep{3dmem}. We build on the broader principle of
persistent scene memory and study its role as an \emph{action interface for
manipulation}. Learned perception writes metric object state; deterministic
association and relation extraction preserve identities and events; a VLM
selects targets from a structured serialization; and frozen low-level skills
execute grounded cues. Our focus is the binding between historical references
and actionable geometry, rather than a new detector, scene mapper, or low-level policy.

We call this system \method (\textbf{O}bje\textbf{c}t-\textbf{C}entric
\textbf{4}D \textbf{M}emory), pronounced ``Occam.'' The name also reflects an
Occam's-razor-inspired factorization: keep persistent state explicit and
low-level control history-free. The representation distinguishes a fixed historical location
from the current location of a persistent object: a vanished place remains
fixed under ego-motion, whereas a selected object's target can follow its live
track during execution.

Our contributions are (1) an object-centric memory interface with world-frame
tracks and explicit temporal, motion, and containment relations; (2) a
record--replay evaluation separating memory decisions from execution, including
a viewpoint control that distinguishes historical retrieval from current-view
grounding; and (3) evidence across seven simulation conditions with one frozen
executor, oracle diagnostics, and a paired real-world memory study. A hardware
perturbation additionally demonstrates live target updates after a single
planning call. Qualitative videos are available at \url{https://occ4m-sup.github.io/occ4m-supplementary/}.

\section{Method}
\label{sec:method}

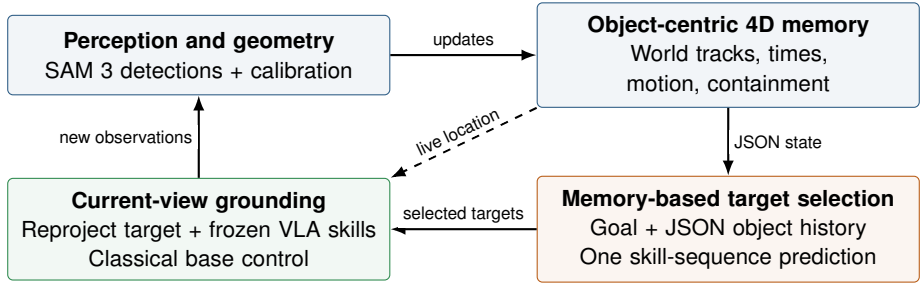
\begin{figure}[t]
\centering
\begin{tikzpicture}[
  box/.style={draw,rounded corners=2pt,align=center,text width=4.7cm,
              minimum height=1.05cm,font=\small,inner sep=5pt},
  arr/.style={-{Latex[length=2mm]},thick},
  lab/.style={font=\scriptsize,fill=white,inner sep=2pt}]
\node[box,draw=memcolor,fill=memcolor!6] (perception) at (0,0)
  {\textbf{Perception and geometry}\\SAM 3 detections + calibration};
\node[box,draw=memcolor,fill=memcolor!6] (memory) at (7,0)
  {\textbf{Object-centric 4D memory}\\World tracks, times, motion, containment};
\node[box,draw=plancolor,fill=plancolor!6] (planner) at (7,-2.3)
  {\textbf{Memory-based target selection}\\Goal + JSON object history\\One skill-sequence prediction};
\node[box,draw=vlacolor,fill=vlacolor!6] (control) at (0,-2.3)
  {\textbf{Current-view grounding}\\Reproject target + frozen VLA skills\\Classical base control};
\draw[arr] (perception) -- node[lab,above]{updates} (memory);
\draw[arr] (memory) -- node[lab,right]{JSON state} (planner);
\draw[arr] (planner) -- node[lab,above]{selected targets} (control);
\draw[arr,dashed] (memory.south west) -- node[lab,sloped,above]{live location} (control.north east);
\draw[arr] (control) -- node[lab,left]{new observations} (perception);
\end{tikzpicture}
\caption{\textbf{Persistent memory as an action interface.} Perception and
calibration update object tracks in a shared frame. A VLM selects identities,
historical locations, and a skill sequence from explicit memory. Execution
reprojects the target into the current view. A historical place stays fixed;
a selected object's location can update from its live track without a new VLM call.}
\label{fig:method}
\end{figure}

Figure~\ref{fig:method} summarizes the system. Each processed RGB observation
updates persistent memory. One VLM call reads the memory at query time and
returns an ordered skill sequence with targets. Perception continues during
execution, while the low-level controller receives only current observations
and a grounded target cue.

\subsection{From pixels to a shared world frame}

For each semantic class, SAM~3 \citep{sam3} is queried with a short noun phrase such as ``red cube'' or ``grey container.'' We convert each detected image location into a world point by intersecting its calibrated camera ray with the horizontal support plane at the corresponding object height, then reject points outside a coarse workspace bound. A target recorded before base motion therefore remains actionable after the viewpoint changes. The current implementation assumes known support geometry; measured depth is an alternative for future deployments. If $\mathbf{x}^{W}$ is a stored world point, its current cue is
\begin{equation}
\mathbf{u}_t=\pi\!\left(\mathbf{K}_t\mathbf{T}_{C_t\leftarrow W}\bar{\mathbf{x}}^{W}\right),
\label{eq:reprojection}
\end{equation}
where $\bar{\mathbf{x}}^{W}$ is homogeneous, $\mathbf{T}_{C_t\leftarrow W}\in\mathbb{R}^{3\times4}$ maps homogeneous world points to camera coordinates, $\mathbf{K}_t$ is the camera intrinsic matrix, and $\pi$ performs perspective division. Ego-motion changes the projection, not the stored historical place.

\subsection{Persistent object tracks}

The memory is a pool of object instances; semantic class is an attribute rather than a dictionary key, allowing multiple same-class objects to coexist. For each frame, detections are matched to tracks within each semantic class by Hungarian assignment on planar world distance, gated at $12\,\mathrm{cm}$ around a constant-velocity prediction. Unmatched detections create tracks. Unmatched tracks are retained, record a miss, and keep their last detected pose; reprojection determines whether their predicted locations lie inside the current field of view. Only a detection updates an object's independent pose estimate. A location inferred through containment is distinct from this independently observed pose. 

Track histories induce task-relevant facts using shared update rules. First-observation time defines appearance order. Motion events retain their start time, pre-motion pose, and current pose. A directed containment edge links an object that disappears beneath a nearby cover to that cover, whose pose provides the object's inferred location until reappearance. Given detections, camera poses, and fixed thresholds, all updates are deterministic. Exact thresholds are summarized in Appendix~\ref{app:implementation}.

\subsection{Memory-to-action interface}

The planner receives a plain-text JSON serialization of the typed graph. Each
track stores a persistent identifier, semantic class, world position,
visibility state, first and last observation times, motion events with
pre-motion and current positions, and containment edges. Gemini~2.5 Flash
\citep{gemini25} is called once per episode. Its prompt asks it to select relevant object identities and copy coordinates
already present in memory, rather than estimate metric geometry itself, and to
return an ordered plan over \{navigate, pick, place, drop, reset\}. Because planned targets are resolved against persistent
object tracks, a selected target can remain bound to an object identity while
its position is updated online. In closed-loop execution, perception continues
to update the corresponding track and the executor cue is re-grounded to its
latest location without re-querying the VLM. Thus, the planner decides
\emph{which} object once, while memory maintains \emph{where} it is. Historical-location targets instead retain their recorded coordinates: they do not follow later motion of the originally observed object.

Execution is factored and the low-level policy remains history-free. A
classical controller moves the mobile base toward each selected world-frame
target. Pick and place use a \emph{single} frozen $\pi_{0.5}$ low-rank adaptation (LoRA) executor
\citep{pi05} for all simulation tasks. The target is reprojected into the
current shoulder image and rendered using the visual marker used during
fine-tuning. The policy receives only current shoulder and wrist images,
joint state, one skill word, and the marker; it does not observe the task
instruction, memory, or prior frames. The unified executor was fine-tuned for
5{,}000 steps on 1{,}100 atomic demonstrations (700 pick and 400 place); training details appear in Appendix~\ref{app:implementation}.

\begin{table}[t]
    \centering
    \caption{Six task families and the viewpoint-transfer variant. T1c combines a new camera pose with a current-frame grounding requirement.}
    \label{tab:tasks}
    \footnotesize
    \setlength{\tabcolsep}{4pt}
    \renewcommand{\arraystretch}{1.08}
    \begin{tabularx}{\textwidth}{@{}c >{\raggedright\arraybackslash}p{0.22\textwidth} >{\raggedright\arraybackslash}X@{}}
        \toprule
        & \textbf{Capability} & \textbf{Required decision} \\
        \midrule
        T1 & Spatial persistence & Place the red cube at the vanished green cube's last location. \\
        T1c & Viewpoint transfer & After the target vanishes and the base moves $0.30\,\mathrm{m}$, express its location in the current view. \\
        T2 & Temporal identity & Place the blue cube at the location of the marker that appeared first, and the red cube at the location of the marker that appeared second. \\
        T3 & Event history & Three cubes move sequentially; pick the second mover and place it at the first mover's pre-motion pose. \\
        T4 & Relational persistence & After identical covers are shuffled, pick the cover containing the red cube. \\
        T5 & Cross-view association & Retrieve a vanished sign location after driving to a second table and back. \\
        T6 & Multi-instance identity & After all cubes become blue, sort them by their initial red/green identity. \\
        \bottomrule
    \end{tabularx}
\end{table}

\section{Experiments}
\label{sec:experiments}

\subsection{Task suite and evaluation protocol}

We evaluate a single-arm mobile Panda in ManiSkill3~\citep{maniskill3}, using egocentric shoulder and wrist RGB cameras. Each episode has a scripted observation phase followed by a manipulation query requiring historical information. We evaluate seven conditions spanning six task families (Table~\ref{tab:tasks}) on the same $N=50$ evaluation seeds. Each observation phase is recorded once and replayed to every method, providing identical RGB histories and goals; the same recorded camera poses are used for geometric conversion. Qualitative sequences for all task families are provided in Appendix~\ref{app:qualitative}.

Before robot motion, each method produces a \emph{memory decision}: required object identities and pick/place target locations, or all required cube assignments for T6. We call correctness of this complete perception--memory--VLM decision \emph{memory success}; it includes target selection and output completeness, not only storage fidelity. Memory and end-to-end success are not nested. Memory success uses a uniform $5\,\mathrm{cm}$ geometric target-error threshold, where applicable, together with task-specific identity constraints, whereas end-to-end success uses each environment's native task criterion. The end-to-end criteria are task-dependent and generally looser: positional tolerances are $8\,\mathrm{cm}$ for T1/T1c and T3, and $12\,\mathrm{cm}$ for T2, T5, and T6, together with the corresponding identity and resting-height requirements; T4 instead succeeds if the robot grasps the correct container. Appendix~\ref{app:vlm_eval} gives the formal metric, frame constraints, and output schema.

All simulation rows using learned actions share the same frozen executor described in Section~\ref{sec:method}. \method uses Gemini~2.5 Flash as its planner, while \framesamp uses Gemini~3.7 Flash \citep{gemini37}. The comparison holds the simulation executor fixed but does not match the VLM backbone or input representation. We call this raw-history baseline \framesamp: the VLM reasons directly over sampled observation frames. The principal FrameSamp configuration receives the complete recorded history with a frame budget of $K=32$ and returns a pixel and frame identifier for each target. Histories contain 10--20 frames except T3 ($\sim$30), making $K=32$ sufficient to cover the complete history. We additionally evaluate $K=16$ uniformly sampled observations as a context-size ablation. The configurations use independent VLM calls and rollouts, so results may differ even when both budgets cover the same full history. By default, \framesamp may ground a target in any supplied historical frame, after which the selected pixel is back-projected using that frame's calibration. This avoids requiring the VLM to express a location that may no longer be visible in the current image and provides calibrated geometric conversion after frame and pixel selection. 

Task semantics impose additional frame constraints. In T4, the pick must be localized in the final frame so that success requires identifying the correct cover after the shuffle. T1c intentionally changes the grounding interface to isolate viewpoint transfer: after the target disappears and the base translates $0.30\,\mathrm{m}$, the remembered location must be expressed in the current post-motion frame. T1b separately controls for this current-frame requirement without ego-motion. The T5 place target is grounded in the original Table-A observation. 

Oracle actions replace learned low-level execution with motion planning. Oracle memory replaces perceptual memory with simulator state while retaining the VLM planner and learned executor. Oracle memory+plan additionally replaces the VLM plan with the task-specific ground-truth plan; it is an oracle diagnostic, not a competing method.

\subsection{Viewpoint-transfer design}
\label{sec:viewpoint_design}

T1 permits \framesamp to point directly to an earlier image in which the vanished green cube is visible. To separate historical recall from viewpoint transfer, T1b requires the place pixel in the current frame but leaves the camera fixed. T1c then translates the base $0.30\,\mathrm{m}$ after the green cube vanishes while retaining the current-frame requirement. The two changes are jointly necessary: T1b tests whether the output constraint alone causes failure, while base motion without the constraint could be bypassed by selecting an earlier frame and using that frame's calibration.

\subsection{Real-world evaluation}\label{sec:real_robot}

We evaluate \method using a stationary Franka arm in a tabletop setup using a fixed camera and a
separate $\pi_{0.5}$ pick policy fine-tuned for 5,000 steps on only 54
pick demonstrations. Because the camera does not move, image coordinates
themselves provide a stable planar reference frame. We therefore store object
locations in a normalized image-plane coordinate system rather than
back-projecting them into a shared 3D world frame. The hardware system retains the same factorization into persistent tracking, containment reasoning, target selection, and execution. It does not test world-frame viewpoint invariance.

We evaluate a compositional memory task over $N=20$ randomized episodes.
A white mug is first observed and removed; three cups then cover colored
cubes and are shuffled. The robot is subsequently instructed to
(1) pick the cup covering the red cube and, after reset,
(2) pick the cup occupying the white mug's original location.

For memory evaluation, the same recorded observation history is replayed to
\method and \framesamp, yielding a paired comparison without differences in
physical trajectories. We restrict the paired hardware comparison to memory
because physical execution would induce method-dependent trajectories, and the
shuffled scene cannot be replayed exactly across rollouts. \framesamp uses
Gemini~3.7 Flash. Memory success is reported for each subtask; joint memory
success requires both predictions to be correct in the same episode. For
\method's physical execution, end-to-end success requires both physical
subtasks to succeed.

We evaluate FrameSamp with \(K\in\{16,32\}\) observations uniformly sampled over each
recorded history, as well as with the complete observation history (``all frames''). \method instead builds persistent state incrementally and queries the VLM only with that state.

\paragraph{Closed-loop perturbation.} We additionally demonstrate identity-bound closed-loop grounding. After
the VLM issues a plan, the selected target is bound to a persistent track updated from live observations during execution. At each $0.8\,\mathrm{s}$ action-chunk boundary,
the cue is re-grounded to the track's latest position. We test this by translating the target cup by approximately $20\,\mathrm{cm}$ while the robot is approaching it.

\section{Results}
\label{sec:results}

\begin{table}[H]
    \centering
    \caption{Memory and end-to-end success (\%) on $N=50$ evaluation episodes per condition. Means average the seven task-condition rates. All learned-action rows use the same frozen executor. The principal raw-history baseline receives the complete history (budget $K=32$).}
    \label{tab:results}
    \footnotesize
    \setlength{\tabcolsep}{4pt}
    \renewcommand{\arraystretch}{1.07}
    \begin{tabular}{@{}lrrrrrrrr@{}}
        \toprule
        \textbf{Method} & \textbf{T1} & \textbf{T1c} & \textbf{T2} & \textbf{T3} & \textbf{T4} & \textbf{T5} & \textbf{T6} & \textbf{Mean} \\
        \midrule
        \multicolumn{9}{@{}l}{\emph{A. Memory success}} \\
        \method & 100 & 100 & 92 & 98 & 88 & 98 & 100 & \textbf{96.6} \\
        \method + oracle actions & 100 & 100 & 92 & 98 & 88 & 98 & 100 & 96.6 \\
        \method + oracle memory & 100 & 100 & 100 & 100 & 100 & 100 & 100 & 100.0 \\
        \method + oracle memory+plan & 100 & 100 & 100 & 100 & 100 & 100 & 100 & 100.0 \\
        \framesamp ($K{=}16$, Gemini~3.7) & 100 & 6 & 72 & 32 & 54 & 100 & 2 & 52.3 \\
        \framesamp ($K=32$, Gemini 3.7) & 100 & 0 & 80 & 28 & 74 & 100 & 0 & \textbf{54.6} \\
        \midrule
        \multicolumn{9}{@{}l}{\emph{B. End-to-end success}} \\
        \method & 94 & 98 & 92 & 90 & 90 & 96 & 62 & \textbf{88.9} \\
        \method + oracle actions & 100 & 100 & 92 & 98 & 92 & 98 & 100 & 97.1 \\
        \method + oracle memory & 98 & 92 & 98 & 92 & 100 & 96 & 76 & 93.1 \\
        \method + oracle memory+plan & 96 & 92 & 100 & 94 & 98 & 96 & 70 & 92.3 \\
        \framesamp ($K{=}16$, Gemini~3.7) & 96 & 6 & 100 & 40 & 54 & 98 & 0 & 56.3 \\
        \framesamp ($K=32$, Gemini 3.7) & 92 & 4 & 100 & 36 & 74 & 98 & 0 & \textbf{57.7} \\
        \bottomrule
    \end{tabular}
\end{table}

\subsection{Main comparison}

Across the seven conditions, \method achieves 96.6\% memory success and
88.9\% end-to-end success, compared with 54.6\% and 57.7\% for
\framesamp given the full $K=32$ observation history
(Table~\ref{tab:results}). The raw-history baseline is competitive
when the relevant state can be recovered directly from the recorded
observations: it reaches 100\% memory success on T1 and T5 and 80\% on T2.
Substantial gaps remain on event-order reasoning (T3), persistent
multi-instance bookkeeping after recoloring (T6), and viewpoint transfer
(T1c); T4 retains a 14-point memory gap even with the complete shuffle history.

Increasing FrameSamp's budget from $K=16$ to the full $K=32$ history
raises mean memory success from 52.3\% to 54.6\% and end-to-end success
from 56.3\% to 57.7\%. The clearest gain is in T4 (54\% to 74\%
memory success), consistent with the value of intermediate shuffle observations.
Denser sampling does not close the gaps on T3, T6, or T1c.

\paragraph{Memory-conditioned policy baseline.} 
We separately train FrameSamp+Modul, the best-performing MME-VLA variant reported in RoboMME~\citep{robomme}, which conditions $\pi_{0.5}$ on sampled history through learned modulation. Its best observed mean success across training sweeps is 9.3\% on T1--T6
(excluding T1c), versus 87.3\% for \method (Appendix~\ref{app:memcond}).

\subsection{Ego-motion-invariant recall}

\begin{figure}[H]
    \centering
    \includegraphics[width=\textwidth]{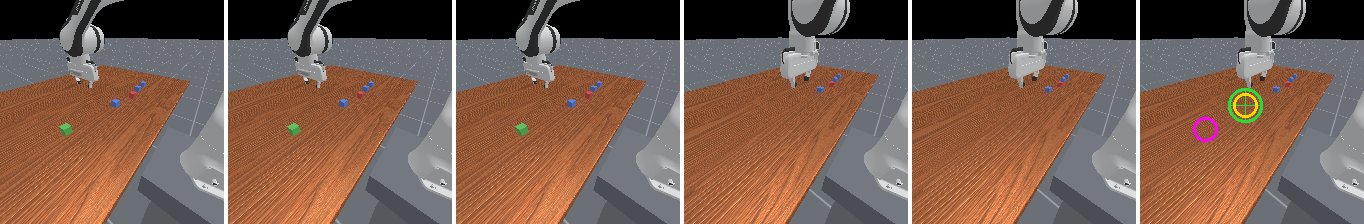}
    \caption{\textbf{T1c viewpoint-transfer example.} The green cube disappears, the base translates $0.30\,\mathrm{m}$, and the robot must place the red cube at the green cube's remembered world location from the new viewpoint. Yellow ring: \method prediction; pink ring: \framesamp (Gemini~3.7 Flash) prediction; green ring: ground-truth target.}
    \label{fig:e1c}
\end{figure}

\begin{table}[H]
    \centering
    \caption{Viewpoint-transfer control. Entries report memory / end-to-end success (\%). T1b changes only the required output frame; T1c additionally moves the base after the target vanishes. \framesamp uses Gemini~3.7 Flash with $K=16$, which covers the full history for T1/T1b/T1c.}
    \label{tab:viewpoint}
    \small
    \setlength{\tabcolsep}{5pt}
    \renewcommand{\arraystretch}{1.12}
    \begin{tabularx}{\textwidth}{@{}Xcc@{}}
        \toprule
        \textbf{Condition} & \textbf{\framesamp} & \textbf{\method} \\
        \midrule
        T1: historical-frame grounding ($N=50$) & 100 / 96 & 100 / 94 \\
        T1b: current frame, fixed viewpoint ($N=50$) & 100 / 92 & 100 / 94 \\
        \textbf{T1c: current frame + $0.30\,\mathrm{m}$ base motion ($N=50$)} & \textbf{6 / 6} & \textbf{100 / 98} \\
        \bottomrule
    \end{tabularx}
\end{table}

Figure~\ref{fig:e1c} illustrates the viewpoint-transfer setup, and Table~\ref{tab:viewpoint} reports the corresponding control results. T1, T1b, and T1c together separate historical retrieval from viewpoint transfer. \framesamp succeeds when it may point to the historical observation containing the target (T1), and remains at 100\% memory success when only the current-frame output constraint is imposed without changing viewpoint (T1b). Under the viewpoint change in T1c, however, \framesamp falls to 6\% memory and end-to-end success, while \method remains at 100\% and 98\%. In the paired memory comparison, \method is correct on all 50 T1c episodes:
47 are \method-only successes and 3 are successes for both methods; there are no \framesamp-only successes. Median error is $1.2\,\mathrm{cm}$ for \method and $28.8\,\mathrm{cm}$ for \framesamp, nearly matching the imposed $30\,\mathrm{cm}$ displacement. This error pattern is consistent with re-pointing at a stale image location rather than transforming the remembered place into the new view. The stored world target remains unchanged by ego-motion. 

\subsection{Error decomposition}

Oracle actions provide the largest observed gain: $29/350$ additional
successes, or 8.3 percentage points, raising end-to-end success to 97.1\%.
Oracle memory adds $15/350$ successes (4.3 points); replacing its VLM plan
with the ground-truth plan changes success by $-3/350$ ($-0.9$ points).
These variants use separate rollouts and, where applicable, new VLM calls;
they are aggregate diagnostics, not perfectly paired causal interventions.
Their small non-monotonic differences should not be interpreted as a cost
of correct planning. The larger oracle-action gain points to execution as
an important remaining bottleneck.

T6 makes this separation especially clear: \method has 100\% memory success
but 62\% end-to-end success, and oracle actions reach 100\%. The learned
executor sorts 90\% of required cubes individually; under an
independence approximation, compounding across four cycles predicts $0.90^4\!\approx\!0.66$, close to the observed episode success. All 19 \method failures on T6 occur after a usable memory/plan
decision. By contrast, \framesamp frequently returns incomplete multi-step
plans on T6, failing to produce a complete one-to-one assignment over all
required objects. This indicates a plan-completeness/interface failure rather
than simple localization error.

\subsection{Real-world evaluation}

\begin{figure}[H]
    \centering
    \includegraphics[width=\textwidth]{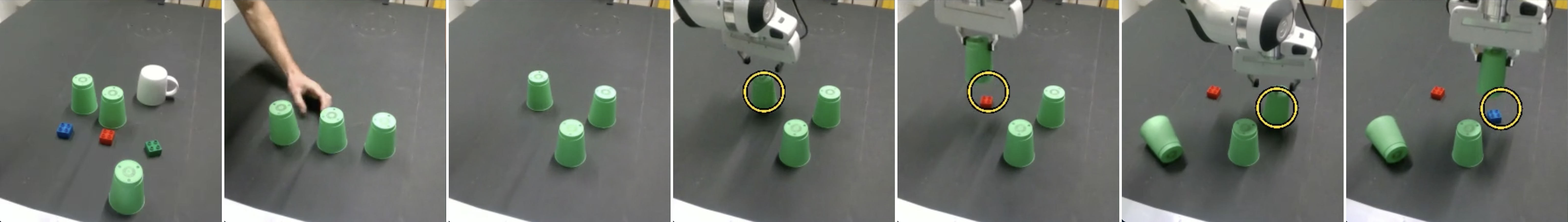}
    \caption{\textbf{Real-world compositional evaluation.} The robot must first
    identify and pick the cup covering the red cube after the cups are shuffled, and then pick the cup occupying the white mug's remembered initial location. The yellow rings are rendered by \method into the visual input of the low-level policy to indicate the target.}
    \label{fig:realworld1}
\end{figure}

\begin{figure}[H]
    \centering
    \includegraphics[width=\textwidth]{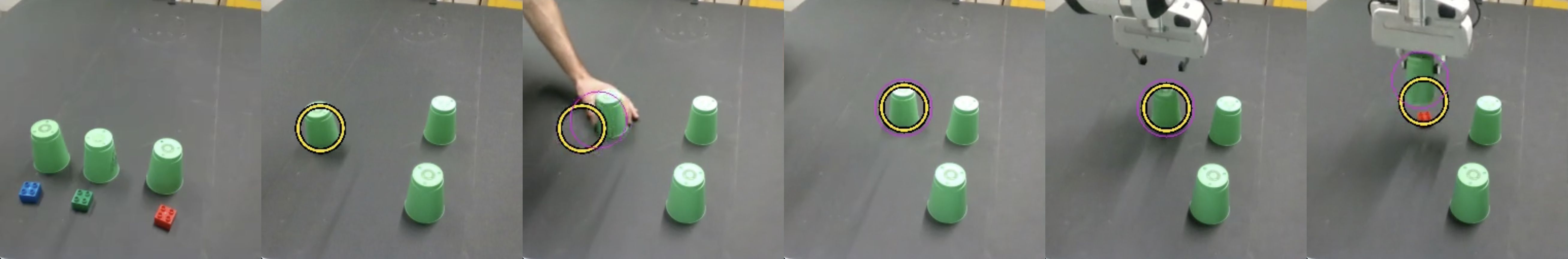}
    \caption{\textbf{Closed-loop identity-bound grounding under perturbation.} After planning, the target cup is moved during execution. \method preserves its identity and updates its location without re-querying the VLM, enabling a successful grasp at the new location. Magenta: memory estimate (not policy input); yellow: cue sent to $\pi_{0.5}$ at each action chunk.}
    \label{fig:realworld2}
\end{figure}

\begin{table}[H]
\centering
\caption{\textbf{Real-world compositional manipulation.}
Memory accuracy measures selection of the correct target cup for each
subtask. S1 denotes shuffled-containment recall and S2 denotes recall of the white
mug's original location; ``Both'' requires both memory decisions to be correct in the same episode. Physical execution success is reported only for \method, and ``E2E'' requires successful completion of both subtasks. All methods are evaluated on the same $N=20$ recorded episodes.}
\label{tab:compositional_results}
\small
\begin{tabular}{lccc|ccc}
\toprule
& \multicolumn{3}{c|}{\textbf{Memory accuracy}}
& \multicolumn{3}{c}{\textbf{Physical execution}} \\
\cmidrule(lr){2-4}
\cmidrule(lr){5-7}
\textbf{Method}
& \textbf{S1}
& \textbf{S2}
& \textbf{Both}
& \textbf{S1}
& \textbf{S2}
& \textbf{E2E} \\
\midrule
\textbf{\method}
& 85\%
& 95\%
& \textbf{85\%}
& 75\%
& 55\%
& \textbf{45\%} \\
\framesamp ($K=16$)
& 30\%
& 95\%
& 30\%
& -- & -- & -- \\
\framesamp ($K=32$)
& 25\%
& 100\%
& 25\%
& -- & -- & -- \\
\framesamp (all frames)
& 25\%
& 90\%
& 20\%
& -- & -- & -- \\
\bottomrule
\end{tabular}
\end{table}

Figure~\ref{fig:realworld1} shows the real-world compositional task. On the paired compositional histories, OCC4M achieves 85\% joint memory accuracy, versus 30\%, 25\%, and 20\% for FrameSamp with $K=16$, $K=32$, and all frames, respectively (Table~\ref{tab:compositional_results}). The gap is concentrated in shuffled-containment recall (S1): OCC4M achieves 85\%, while FrameSamp remains at 25--30\%, around the 33.3\% uniform-choice reference even with all frames. All methods perform well on historical-location recall (S2). Joint memory accuracy is 17/20 for \method and 6/20 for the best FrameSamp setting. 

When \method's decisions are executed on the Franka, 15/20 first subtasks
and 11/20 second subtasks succeed, with 9/20 episodes (45\%) completing the
full two-stage sequence. Given that the physical executor is trained from only 54 pick demonstrations,
the gap between 85\% joint memory accuracy and 45\% end-to-end success further
suggests that low-level execution, rather than memory, is the dominant
remaining bottleneck.

\paragraph{Closed-loop grounding under perturbation.}
Figure~\ref{fig:realworld2} shows the closed-loop perturbation demonstration,
in which the target cup is displaced by approximately $20\,\mathrm{cm}$ after
the plan has been issued. The persistent track enables rapid online
retargeting: the selected object identity remains fixed while its location is
updated at $10.6\,\mathrm{Hz}$ without re-querying the VLM. The rendered cue
is refreshed at action-chunk boundaries, implying approximately
$0.5\,\mathrm{s}$ expected retargeting latency under a uniform chunk-phase assumption. During the demonstrated
perturbation, the cue is updated six times with no target re-binding, and the
robot ultimately grasps the cup at its displaced location.

\section{Discussion and Limitations}
\label{sec:limitations}

\paragraph{Historical retrieval and actionable spatial memory.}
T1/T1b/T1c distinguish remembering an earlier observation from expressing its
target after viewpoint change. FrameSamp remains at 100\% memory success
under the current-frame requirement alone, but falls to 0--6\% after base
motion; \method retains 100\%. Because both frame budgets cover the complete history in T1 and T1c,
the viewpoint-transfer failure cannot be explained by omitted observations. Explicit world coordinates preserve the historical place while
its image projection changes, allowing the remembered target to be grounded
from the robot's current viewpoint. Beyond geometric persistence, the gaps
on event-order and containment tasks support explicitly maintaining
task-relevant identities and relations across observations. The resulting
interface supplies the planner with persistent state and actionable
coordinates, reducing the information it must reconstruct from images at
query time. The evidence concerns the evaluated system interfaces:
different VLMs and input representations prevent attributing the entire
gap to memory structure alone. A raw-history system augmented with explicit
geometric state is not tested here.

\paragraph{Physical grounding and execution.}
The hardware study provides initial evidence on paired physical histories:
85\% joint memory accuracy versus at most 30\% for FrameSamp, with 45\%
end-to-end completion for \method. Only \method is executed physically,
and its fixed-camera image-plane memory does not establish mobile world-frame
transfer. The single perturbation demonstration shows live grounding of an
already selected identity; it is not an evaluation of general disturbance
robustness or high-level plan repair. Improved manipulation skills remain
necessary to turn correct spatial decisions into reliable task completion.

\paragraph{Incremental grounding during execution.}
Persistent object identities allow target selection and geometric updates
to operate at different timescales. Once the planner selects an object,
new observations can update its location and refresh the executor's cue
without repeating task-level reasoning. This complements historical-place
recall: a remembered place remains fixed, whereas a selected object's
target follows its live track. The perturbation demonstration illustrates
this distinction, with \method updating tracks at 10.6\,Hz and successfully
grasping the displaced cup without a new VLM call. On hardware, FrameSamp's
mean query latency over 20 histories is 11.7\,s, 12.2\,s, and 31.6\,s for
$K=16$, $K=32$, and all frames (300--600; 497 on average).
These deployment-dependent timings compare local-GPU perception with
cloud VLM queries, rather than establishing an inherent latency advantage.
The architectural benefit is that refreshing an already selected target
does not require resubmitting the visual history to the planner.

Evaluation is limited to small structured scenes and scripted histories.
Known support geometry, fixed association thresholds, and uncalibrated point
estimates limit robustness under clutter, long occlusions, and ambiguous
associations. The memory-conditioned policy sweep does not exclude sensitivity
to other training settings. Matched-backbone and component ablations, uncertainty-aware association, and
real-world mobile viewpoint transfer are the next tests of generality.

\section{Conclusion}

\method makes persistent spatial memory an explicit interface between
perception, task reasoning, and history-free manipulation. Across seven
simulation conditions, it achieves 96.6\% memory and 88.9\% end-to-end
success, versus 54.6\% and 57.7\% for a full-history VLM baseline with
the same executor. The viewpoint controls show the value of preserving a
world-frame place across ego-motion, while the physical experiments provide
initial evidence for relational recall and live target grounding. These results
support persistent object-centric state as a practical basis for embodied
spatiotemporal reasoning and identify execution and association as priorities
for more reliable deployment.

\begin{ack}
This work was partially funded by Office of Naval Research grant N00014-25-1-2322.
\end{ack}

\clearpage
\appendix

\section{Implementation Details}
\label{app:implementation}

\paragraph{Unified executor.}
All learned-action simulation results use a single frozen $\pi_{0.5}$ LoRA checkpoint. It was fine-tuned for 5{,}000 steps on 1{,}100 atomic demonstrations: 700 pick and 400 place demonstrations spanning the task environments. Both \method and the shared-executor \framesamp baseline use this identical checkpoint. The policy receives current shoulder/wrist RGB, joint state, one atomic skill word, and one rendered target cue.

\paragraph{Tracking and control.}
Detections are associated to tracks within each semantic class by Hungarian assignment on planar world distance, gated at $12\,\mathrm{cm}$ around a constant-velocity prediction. Motion onset is recorded when an object departs its rest pose by more than $5\,\mathrm{cm}$; events retain the onset time, pre-motion pose, and current pose. When an object disappears beneath a cover within $11\,\mathrm{cm}$, a containment edge links it to the cover, whose pose provides an inferred location until reappearance. The classical base controller follows a straight-line trajectory with a velocity ramp toward the selected world-frame target. In T6, recoloring leaves poses unchanged, so remembered original-color tracks retain valid target locations; this experiment does not establish cross-appearance re-identification.

\paragraph{Real-world perception timing.}
During closed-loop execution, the tracker processes successive live frames
back-to-back. For the perturbation experiment, a SAM 3 detection call on the warm local server takes
$88\,\mathrm{ms}$ median ($89\,\mathrm{ms}$ p90, $93\,\mathrm{ms}$ maximum), while the subsequent memory-association update takes
$0.04\,\mathrm{ms}$ median. In the reported perturbation episode,
the tracker performs $382$ updates over $36.0\,\mathrm{s}$, corresponding to
$10.6$ updates/s. The executor cue may be refreshed at $0.8\,\mathrm{s}$
action-chunk boundaries. Assuming the displacement occurs uniformly within a chunk, the mean wait to the next boundary is $0.4\,\mathrm{s}$; adding about $0.09\,\mathrm{s}$ for perception gives approximately $0.5\,\mathrm{s}$. A full chunk plus perception gives a nominal bound of approximately $0.9\,\mathrm{s}$. These are scheduling estimates, not measured latency quantiles or hard real-time guarantees.

\section{Memory-Conditioned Policy Baseline}
\label{app:memcond}

We use the FrameSamp+Modul architecture of RoboMME~\citep{robomme},
fine-tuned from the same $\pi_{0.5}$ base model. We train one multi-task policy with 250 demonstrations per original task
(1{,}500 total over T1--T6, excluding T1c) for up to 40{,}000 gradient
steps, evaluating every 10{,}000 steps. At 250 demonstrations per task,
performance reaches 5.0\% mean end-to-end success at 20{,}000 steps and does
not improve with further training. A demonstration-count sweep reaches a best
observed mean success of 9.3\% with 400 demonstrations per task at
20{,}000 steps. Across all tested settings, success remains below 10\%,
compared with 87.3\% for \method on the same six conditions.

\begin{table}[H]
\centering
\caption{Robustness of the multi-task FrameSamp+Modul policy to demonstration
count and training duration. Entries are mean end-to-end success (\%) on the
six original task conditions (T1--T6, excluding T1c).}
\label{tab:framesamp_robustness}
\small
\setlength{\tabcolsep}{8pt}
\begin{tabular}{@{}lcccc@{}}
\toprule
\multicolumn{5}{c}{\textbf{Varying demonstrations per task (20k steps)}} \\
\midrule
Demonstrations & 80 & 250 & 400 & -- \\
Mean success   & 8.3 & 5.0 & 9.3 & -- \\
\midrule
\multicolumn{5}{c}{\textbf{Varying training duration (250 demonstrations/task)}} \\
\midrule
Training steps & 10k & 20k & 30k & 40k \\
Mean success   & 1.7 & 5.0 & 3.3 & 3.3 \\
\bottomrule
\end{tabular}
\end{table}

Success remains below 10\% across these sweeps, but optimizer settings,
architecture adaptation, and demonstration quality remain possible explanations.
The six-task score excludes T1c.

\section{VLM Memory-Baseline Evaluation}
\label{app:vlm_eval}

\paragraph{Memory-success metric.}
For tasks requiring metric pick or place targets, let $\mathcal{T}$ be the set of required target locations and let $\hat{\mathbf{x}}_j,\mathbf{x}^{*}_j\in\mathbb{R}^{2}$ be the predicted and ground-truth planar world coordinates for target $j$. We define
\begin{equation}
    e_{\mathrm{mem}}=\max_{j\in\mathcal{T}}\left\|\hat{\mathbf{x}}_j-\mathbf{x}^{*}_j\right\|_2.
\end{equation}
An episode succeeds if $e_{\mathrm{mem}}<5\,\mathrm{cm}$ and all task-specific identity constraints hold. The maximum is over all required pick and place targets; only pick targets are scored for pick-only tasks.

\begin{table}[H]
\centering
\caption{Task-specific definitions of the executor-independent memory metric.}
\label{tab:memory_metric_definitions}
\footnotesize
\renewcommand{\arraystretch}{1.25}
\setlength{\tabcolsep}{5pt}
\begin{tabularx}{\textwidth}{@{}p{0.06\textwidth}p{0.23\textwidth}p{0.28\textwidth}X@{}}
\toprule
\textbf{Task} & \textbf{Pick target (GT)} & \textbf{Place target (GT)} & \textbf{Additional constraints} \\
\midrule
T1 & red cube (current) & remembered green-cube location & N/A \\
T1c & red cube (current) & same remembered location, expressed after base motion & place prediction uses final frame \\
T2 & blue cube, red cube & blue $\rightarrow$ first marker; red $\rightarrow$ second marker & labels and order correct \\
T3 & second cube that moved (current) & pre-motion location of first mover & pick is second mover \\
T4 & current correct grey cover & N/A (pick-only) & correct cover among three, pick prediction uses final frame \\
T5 & query-color cube on Table B & remembered query-color sign location on Table A & queried color correct \\
T6 & all initially red/green cubes & N/A for memory scoring & one-to-one matching; counts/classes correct \\
\bottomrule
\end{tabularx}
\end{table}

T1b uses the T1 memory metric but requires the place prediction in the final frame, as in T1c; unlike T1c, the camera remains fixed. For T6, predicted and ground-truth instances are matched one-to-one within each original color class using nearest planar distance. Predicted counts must match ground-truth counts before the maximum matched distance is tested. Bin locations are fixed task geometry and excluded from the memory metric.

\paragraph{Frame-indexed grounding.}
Visual observations are non-privileged $224\times224$ RGB frames from \path{panda_shoulder}. Camera intrinsics, world-to-camera extrinsics, and base pose are stored only to convert image-space outputs into world coordinates; they are not shown to the VLM. Image points use normalized $[y,x]\in[0,1000]^2$ coordinates. Each predicted point includes a supplied frame identifier, and the pixel is back-projected using that frame's calibration and the target-height support plane.

The default prompt permits grounding a historical target in any supplied frame. T4 instead requires \texttt{pick\_frame} to be the final frame so the model must track the relevant cover through the shuffle. T1c requires \texttt{place\_frame} to be the final post-motion frame. T5 uses the original Table-A frame for its historical place target. If an episode contains fewer than $K$ observations, all available frames are supplied once.

Each object-transfer prediction has the form
\begin{center}
\small
\texttt{\{pick:[y,x], place:[y,x], pick\_label, pick\_frame, place\_frame\}},
\end{center}
with one cycle per required object transfer. Malformed or empty outputs are not re-prompted and count as memory failures.

\paragraph{Viewpoint-transfer prompt addendum.}
For T1b and T1c, we append the following constraint; in T4, it applies to \texttt{pick\_frame}:
\begin{quote}
\footnotesize\ttfamily\raggedright
You MUST give the place point in the CURRENT (last) frame and set\linebreak
place\_frame = \{cur\}. Use earlier frames only to remember the vanished target;\linebreak
do not point to an earlier frame.
\end{quote}

\clearpage
\section{Qualitative Task Sequences}
\label{app:qualitative}

\begin{figure}[H]
    \centering
    \includegraphics[width=\textwidth]{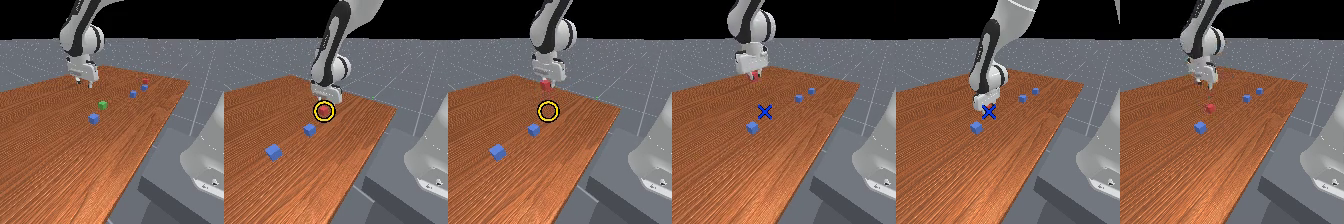}
    \caption{T1: Place the red cube where the green cube was.}
    \label{fig:e1}
\end{figure}

\begin{figure}[H]
    \centering
    \includegraphics[width=\textwidth]{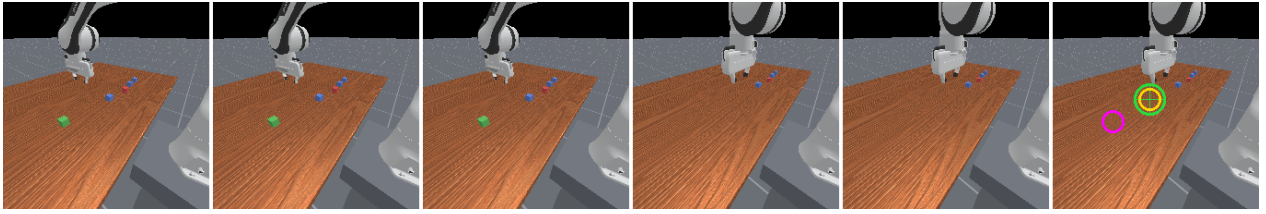}
    \caption{T1c: After the green cube disappears, the mobile base translates $0.30\,\mathrm{m}$, and the robot must place the red cube at the green cube's remembered location from the new viewpoint. Yellow ring: \method prediction; pink ring: \framesamp (Gemini~3.7 Flash) prediction; green ring: ground-truth target.}
    \label{fig:e1c2}
\end{figure}

\begin{figure}[H]
    \centering
    \includegraphics[width=\textwidth]{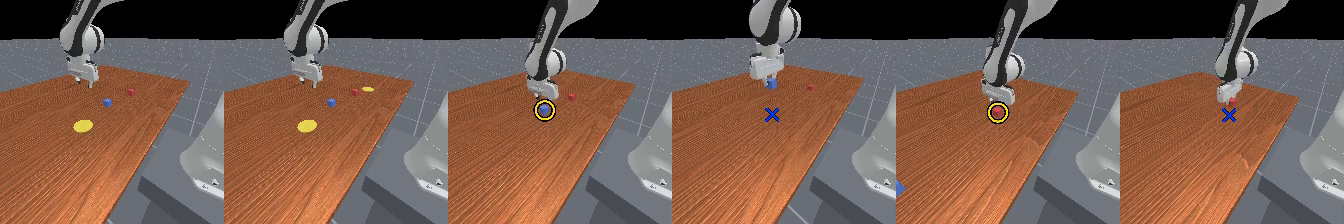}
    \caption{T2: Place the blue cube at the location of the marker that appeared first, and the red cube at the location of the marker that appeared second.}
    \label{fig:e2}
\end{figure}

\begin{figure}[H]
    \centering
    \includegraphics[width=\textwidth]{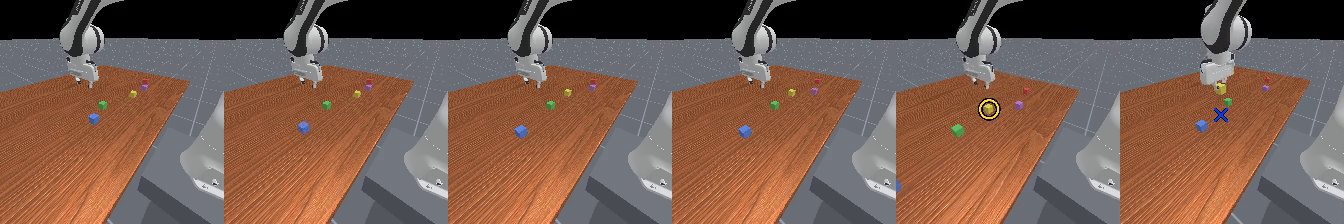}
    \caption{T3: Pick the second mover and place it at the first mover's pre-motion location.}
    \label{fig:e3}
\end{figure}

\clearpage
\begin{figure}[H]
    \centering
    \includegraphics[width=\textwidth]{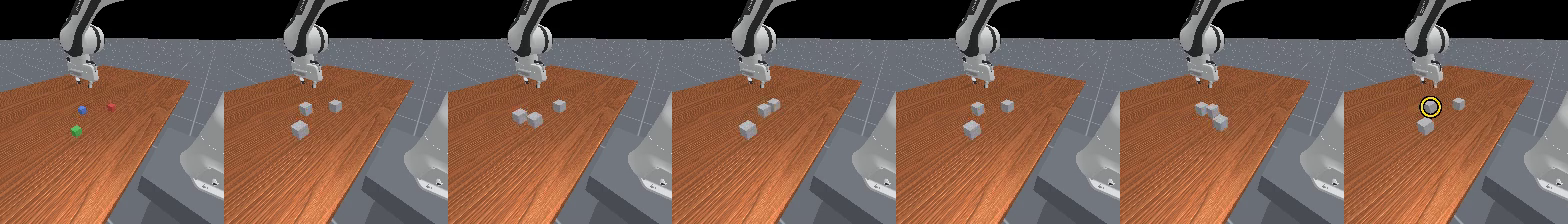}
    \caption{T4: Pick the grey cover containing the red cube after shuffling.}
    \label{fig:e4}
\end{figure}

\begin{figure}[H]
    \centering
    \includegraphics[width=\textwidth]{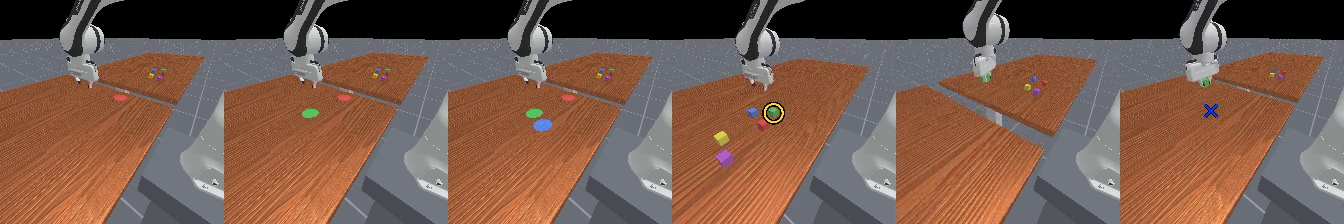}
    \caption{T5: Place the query-color cube at the remembered sign location on the first table.}
    \label{fig:e6}
\end{figure}

\begin{figure}[H]
    \centering
    \includegraphics[width=\textwidth]{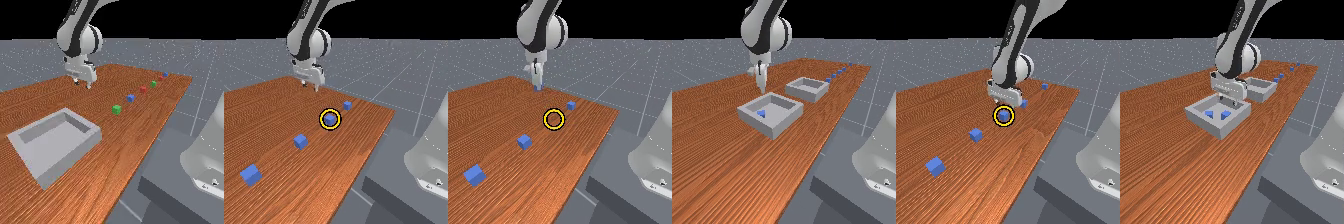}
    \caption{T6: After recoloring, sort the cubes according to their original colors.}
    \label{fig:e8}
\end{figure}

\end{document}